\documentclass{article}

    \usepackage[preprint, nonatbib]{neurips_2020}

\usepackage[utf8]{inputenc} 
\usepackage[T1]{fontenc}    
\usepackage{hyperref}       
\usepackage{url}            
\usepackage{booktabs}       
\usepackage{amsfonts}       
\usepackage{nicefrac}       
\usepackage{microtype}      

\usepackage{tikz}
\usepackage{circuitikz}
\usetikzlibrary{fit,bayesnet,calc}
\usepackage{dsfont}

\usepackage{algorithm}
\usepackage{algorithmic}

\usepackage{amsmath}
\usepackage{graphicx}
\usepackage{xcolor}
\usepackage{gensymb}

\newcommand{\program}{\rho}

\DeclareMathOperator*{\argmin}{arg\,min} 
\DeclareMathOperator*{\argmax}{arg\,max} 
\newcommand{\indicator}{\mathds{1}} 

\usepackage{footnote}
\makesavenoteenv{tabular}

\title{People learn to draw abstract figures by inferring program-like compositional subroutines}

\title{Building new from old: a drawing task to compare how humans and models learn compositional programs}

\title{Building new from old: a drawing task to study how humans and models build new compositional programs compositional programs}

\title{A drawing task to study how humans and models learn new compositional program-like concepts}

\title{Learning abstract structure for drawing by neurally-guided program induction}
\title{Learning abstract structure for drawing by efficient motor program induction}

\author{{Lucas Y. Tian} \\
 Brain and Cognitive Sciences, MIT\\
 \texttt{lyt@mit.edu}
 \And {Kevin Ellis} \\
 Brain and Cognitive Sciences, MIT\\
 \texttt{ellisk@mit.edu}
 \And {Marta Kryven} \\
 Brain and Cognitive Sciences, MIT\\
 \texttt{mkryven@mit.edu}
 \And {Joshua B. Tenenbaum} \\
 Brain and Cognitive Sciences, MIT\\
 \texttt{jbt@mit.edu}}

\begin{document}

\maketitle


\begin{abstract}
Humans flexibly solve new problems that differ qualitatively from those they were trained on. This ability to generalize is supported by learned concepts that capture structure common across different problems. Here we develop a naturalistic drawing task to study how humans rapidly acquire structured prior knowledge. The task requires drawing visual objects that share underlying structure, based on a set of composable geometric rules. We show that people spontaneously learn abstract drawing procedures that support generalization, and propose a model of how learners can discover these reusable drawing programs. Trained in the same setting as humans, and constrained to produce efficient motor actions, this model discovers new drawing routines that transfer to test objects and resemble learned features of human sequences. These results suggest that two principles guiding motor program induction in the model - abstraction (general programs that ignore object-specific details) and compositionality (recombining previously learned programs) - are key for explaining how humans learn structured internal representations that guide flexible reasoning and learning. 
\end{abstract}

\section{Introduction}
A long-term goal of Artificial Intelligence (AI) is to build machines that can quickly learn to solve qualitatively new problems. 
Inspiration may be gained from humans, who readily solve many kinds of tasks without extensive supervision or experience, such as understanding the meaning of new words, 
or rapidly learning new video games \cite{ormrod2016human}. 
These abilities are partly 
supported by learned internal models, or inductive biases,
the structure of which captures regularities useful for reasoning about new situations (e.g., real-world object categories and their relations, or causal rules in video games)~\cite{chi2014nature, harlow1949formation, bartlett1932c, lake2017bbs, lake2019people, tenenbaum2011grow}.  
In this work we adopt a scientific goal, rather than an engineering goal: to probe diagnostic elements of how humans \textit{acquire} structured prior knowledge, and to understand it in computational terms. 

\begin{figure}[ht]
\begin{center}
    \includegraphics[width=0.9\linewidth]{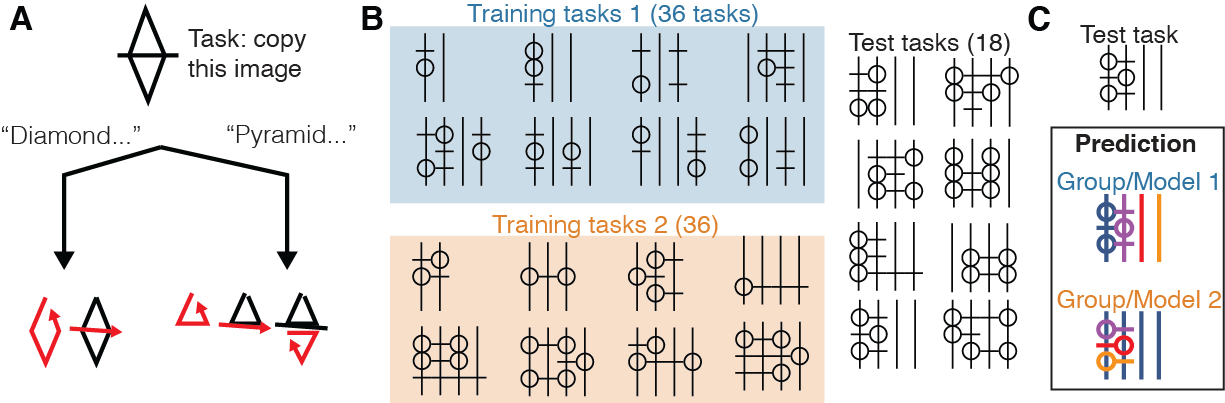}
    \caption{ Background and overview. \textbf{(A)} Prior knowledge influences how people draw. Consider how one might draw the object on top. Drawing tends to differ depending on what structural description is associated with the object - "Diamond with a cross-line" vs. "Pyramid and its reflection"~\cite{van1984drawing}. \textbf{(B)} Representative tasks in Training sets 1 and 2, and in the common Test set. \textbf{(C)} Hypothesized behavior on test tasks that would be diagnostic of learned abstract structure.} 
    \label{background}
\end{center}
\vspace{-6mm}
\end{figure}

To study such learning in a controlled
setting, we introduce a drawing task to probe rapid, few-shot updating of structured internal models after brief experience.
On its surface, the task is simple: to copy, by drawing, a series of novel visual objects. However, \textit{how} people draw provides rich insight into their internal representations and prior knowledge. 
This is intuitive when one considers the inherent ambiguity in how even simple line-drawings should be copied. How do different line segments group into coherent objects? How are those objects related? Drawing is therefore a window into how reasoning and problem-solving is guided by structured prior knowledge, including concepts as diverse as geometry, real-world objects, and geological formations \cite{van1984drawing, karmiloff1990constraints, forbus2011cogsketch, bartlett1932c, cheng2001drawing, fan2018common, goodnow1977children}. For example, consider copying the object in Figure~\ref{background}A. The order of strokes people use to copy this object depends whether it is described as a ``diamond with a cross-line'' or an ``Egyptian pyramid and its reflection on water''  (Figure~\ref{background}A) \cite{van1984drawing}. 

How prior knowledge guiding drawing is learned has been the subject of descriptive studies of drawing behavior across children and adults (e.g., \cite{goodnow1977children, van1984drawing, karmiloff1990constraints, long2018drawings}). Other studies have described single-session learning of specific objects, such as cubes or prisms (e.g., \cite{phillips1985discovery, pemberton1987drawing}). Our goal is to extend prior studies by combining a focus on learning that is rapid and generalizable with a formal computational account of this rapid learning.  

Motivated by prior empirical and theoretical studies (and supported \textit{post hoc} by behavior in this study),
we model learning by incorporating two key principles: 
\textit{abstraction} and \textit{compositionality}. Abstract refers to higher-order structure that is independent of object-specific features. In principle, this supports reasoning that is flexible even in novel situations that differ in lower-level features; e.g., the concept of \textit{repeat} can apply to any simple drawn object \cite{tenenbaum2011grow, lazaro2019beyond, harlow1949formation,kemp2008discovery,jaekel2009grammar}. Compositional refers to complex concepts learned by combining simpler conceptual building blocks. E.g., \textit{repeat} and \textit{hexagon} can be combined to draw an object along the perimeter of an imaginary hexagon. Compositionality gives complexity and variation that extrapolates beyond direct training experience \cite{lake2017bbs, fodor1988connectionism, tenenbaum2011grow, lake2019compositional, lazaro2019beyond}.
Our model realizes abstraction and compositionality through \emph{program induction}.
Concretely, it learns to synthesize abstract, compositional graphics programs from the same visual stimuli given to human subjects,
drawing on recent neuro-symbolic program induction algorithms~\cite{ellis2018learning,dc2020}.
Our behavioral data provides evidence that humans indeed perform few-shot updating of their inductive biases by learning program-like drawing procedures that guide generalization. We describe a program-induction algorithm 
that discovers new abstract, compositional, drawing routines given the same limited training data given to humans. Moreover, the model's learned drawing behavior captures certain diagnostic features of how humans generalize. These results suggest that abstraction and compositionality are key principles for explaining how humans rapidly learn program-like structure that guides reasoning and planning in drawing.

\section{A neuro-symbolic model for learning compositional drawing programs}
Our model is inspired by prior approaches modeling handwriting and drawing \cite{lake2015human, TikZ, forbus2011cogsketch}, AI models that learn structured inductive biases in other cognitive domains \cite{cheyette2017knowledge, rule2018learning, lazaro2019beyond, wingate2013compositional}, and models of perception as inference of symbolic descriptions \cite{erdogan2015sensory, jaekel2009grammar, rock1983logic, jaekel2009grammar, yildirim2019efficient, romano2017bayesian}. We model drawing behavior based on programs, or symbolic procedures representing a description of a drawing's parts--here, simple primitives such as lines and circles--and higher-order relations - e.g., repetition and hierarchy. For a given image, the model infers how to draw it by performing probabilistic inference over a space of drawing programs
. Learning works by estimating a prior over programs in a hierarchical Bayesian fashion. 
Estimating this prior involves inducing new drawing subroutines that are useful for multiple drawing tasks, effectively caching and reusing motor program schemas.
These learned subroutines are \textit{abstract} and \textit{compositional}.
Finally, for comparison with human behavior, we convert these programs into low-level motor actions.
This conversion rests on a third principle relevant for action planning--\textit{motor efficiency}--which is formalized by modeling a proxy for motor costs.
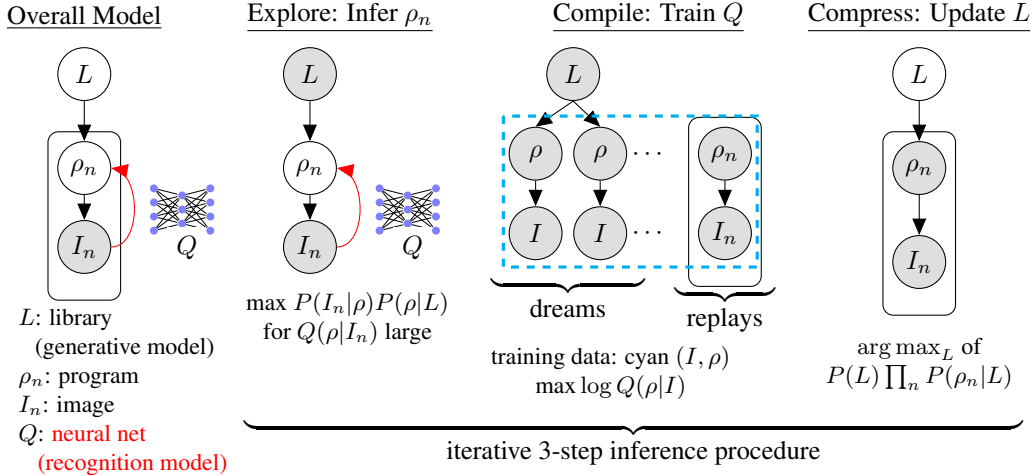
\begin{figure}[h]
\centering
\begin{tikzpicture}
  \begin{scope}[shift={(-7,0)}]
    \node at (3.5,3.75) {\underline{Overall Model}};
    \node[latent] at (3.5,3) (dx){$L$};
    \node[latent] at (3.5,1.75) (zp){$\rho_n$};
    \node[obs] at (3.5,0.7) (xp) {$I_n$};
    \edge{zp}{xp}
    \plate {}{(zp)(xp)}{};
    \draw [->,red] (xp.east) to[out = 0,in = -10] node(nn){} (zp.east);
    \draw [->] (dx) -- (zp);
    \node at (nn) {
      \begin{tikzpicture}[x=2.5cm,y=1.25cm,transform canvas={scale=0.15,shift={+(0.7,2.5)}}]
        \tikzstyle{neuron}=[circle,fill=blue!50,minimum size=20pt]
        \fill[fill=white] (-0.25,-0.5) rectangle (2.25,-4.5);
        \node[rectangle] at (1,1) {};
        \foreach \name / \y in {1,...,4}
            \node[neuron] (I-\name) at (0,-\y) {};
        \foreach \name / \y in {1,...,3}
            \node[neuron] (H-\name) at (1,-\y-0.5) {};
        \foreach \name / \y in {1,...,4}
            \node[neuron] (O-\name) at (2,-\y) {};
        \foreach \source in {1,...,4}
            \foreach \dest in {1,...,3}
                \draw [-latex] (I-\source) -- (H-\dest);
        \foreach \source in {1,...,3}
            \foreach \dest in {1,...,4}
                \draw [-latex] (H-\source) -- (O-\dest);
      \end{tikzpicture}
    };
    \node[shift={+(0.7,-0.5)}] at (nn) [fill=white]{ $Q$ };

    \node[align=left,anchor=north west] at (2.5,0) {\small $L$: library\\\small\phantom{$L$}(generative model)\\\small $\rho_n$: program\\\small $I_n$: image\\\small $Q$: \color{red}{neural net}\\\small\phantom{$Q$}\color{red}{(recognition model)}};
  \end{scope}
  \begin{scope}[shift={(-4,0)}]
    \node at (3.9,3.75) {\underline{Explore: Infer $\rho_n$}};
    \node[obs] at (3.5,3) (dx){$L$};
    \node[latent] at (3.5,1.75) (zp){$\rho_n$};
    \node[obs] at (3.5,0.7) (xp) {$I_n$};
    \node[align = center] at ([yshift = -0.6cm,xshift = 0.5cm]xp.south) {%
      \small$\text{max }P(I_n|\rho)P(\rho|L)$\\\small for $Q(\rho|I_n)\text{ large}$};
    \edge{zp}{xp}
    \draw [->,red] (xp.east) to[out = 0,in = -10] node(nn){} (zp.east);
    \draw [->] (dx) -- (zp);
    \node at (nn) {
      \begin{tikzpicture}[x=2.5cm,y=1.25cm,transform canvas={scale=0.15,shift={+(0.7,2.5)}}]
        \tikzstyle{neuron}=[circle,fill=blue!50,minimum size=20pt]
        \fill[fill=white] (-0.25,-0.5) rectangle (2.25,-4.5);
        \node[rectangle] at (1,1) {};
        \foreach \name / \y in {1,...,4}
            \node[neuron] (I-\name) at (0,-\y) {};
        \foreach \name / \y in {1,...,3}
            \node[neuron] (H-\name) at (1,-\y-0.5) {};
        \foreach \name / \y in {1,...,4}
            \node[neuron] (O-\name) at (2,-\y) {};
        \foreach \source in {1,...,4}
            \foreach \dest in {1,...,3}
                \draw [-latex] (I-\source) -- (H-\dest);
        \foreach \source in {1,...,3}
            \foreach \dest in {1,...,4}
                \draw [-latex] (H-\source) -- (O-\dest);
      \end{tikzpicture}
    };
    \node[shift={+(0.7,-0.5)}] at (nn) [fill=white]{ $Q$ };
  \end{scope}
  \begin{scope}[shift={(0.0,0)}]
    \node at (4,3.75) {\underline{Compile: Train $Q$}};
    \node[obs] at (3,3) (dt){$L$};
    \node[obs] at ([yshift = -0.7cm,xshift = -0.5cm]dt.south) (p2){$\rho$};
    \node[obs] at ([yshift = -0.7cm,xshift = 0.cm]p2.south) (x2){$I$};
    \draw [->] (p2.south) -- (x2.north);
    \draw [->] (dt.south) -- (p2.north);
    \node[obs] at ([yshift = 0cm,xshift = 0.5cm]p2.east) (p3){$\rho$};
    \node[obs] at ([yshift = -0.7cm,xshift = 0.0cm]p3.south) (x3){$I$};
    \draw [->] (p3.south) -- (x3.north);
    \draw [->] (dt.south) -- (p3.north);
    \node at ([yshift = 0cm,xshift = 0.3cm]p3.east) {$\cdots $};
    \node at ([yshift = 0cm,xshift = 0.3cm]x3.east) {$\cdots $};
      
    \node[obs] at ([yshift = 0cm,xshift = 1.3cm]p3.east) (zp){$\rho_n$};
    \node[obs] at ([yshift = 0cm,xshift = 1.3cm]x3.east) (xp) {$I_n$};
    \draw [->] (zp.south) -- (xp.north);
    \plate {}{(zp)(xp)}{};
    \draw[dashed,cyan,very thick] ([yshift = 0.5cm,xshift = -2pt]p2.west)
    rectangle  ([yshift = -0.45cm,xshift = +2pt]xp.east);
    \node at ($(x3)!0.5!(x2) + (0,-0.75)$) {$\underbrace{\phantom{sttesttesttest}}_{\text{\normalsize dreams}}$};
    \node at ($(xp.south) + (0,-0.55)$) {$\underbrace{\phantom{sttesttes}}_{\text{\normalsize replays}}$};
    \node[align = center] at ([yshift = -1.5cm,xshift = 1cm]x2.south) {\small training data: cyan $(I,\rho)$\\
    \small$\text{max}\log Q(\rho|I)$};
  \end{scope}    
  \begin{scope}[shift={(6.6,0)}]
    \node at (1,3.75) {\underline{Compress: Update $L$}};
    \node[latent] at (1,3) (d){$L$};
    \node[obs] at (1,1.75) (z){$\rho_n$};
    \node[obs] at (1,0.5) (x) {$I_n$};
    \edge {z}{x};
    \edge {d}{z};
    \plate {}{(z)(x)}{};
    \node[align = center] at ([yshift = -1cm]x.south) {
    \small$\argmax_L$ of\\\small$P(L)\prod_nP(\rho_n|L)$};
  \end{scope}
  \node at (3.8,-1.8) {$\underbrace{\phantom{testtesttesttesttesttestesttesttesttesttesttesttesttesttesttesttesttest}}_{\text{\normalsize iterative 3-step inference procedure}}$};
  \label{algo}
\end{tikzpicture}
\caption{The Bayesian neurosymbolic program induction algorithm that underlies our computational model, based on~\cite{dc2020,ecc}. Left (``overall model''): Each observed image $I_n$ is explained using a latent program $\rho_n$. The prior or inductive bias is modeled by an inventory or ``library'' of learned primitives, $L$. A neural network recognition model (red arrows) learns to map from images to a distribution over source code of programs likely to explain that image. Conditional distribution output by the network is notated $Q(\cdot |\cdot )$. Inference iterates through \textbf{Explore}, which searches programs ordered under $Q$ and rescores them under true posterior $P(\cdot |L,I_n)$; \textbf{Compile}, which trains the neural network $Q$ to search for image-explaining programs, training both on replays of programs from Explore and ``dreams,'' or samples from the learned prior; and \textbf{Compress}, which updates the prior by compressing out new compositional code abstractions which are incorporated into the library $L$.}
\end{figure}



\paragraph{Program-Induction Model} We treat drawing as Bayesian inference over the most likely program $\program$ that generated each image. Programs are sampled from a generative model defined by a library of primitives $L$ (models are initialized with the library in Table \ref{t:primitives}) .
The model recovers  $\program$ maximizing:
\begin{equation}\label{programMap}
  P(\program|I,L) \propto \underbrace{P(I|\program)}_{\text{likelihood: }1\left[\program\text{ draws }I \right]}\times\underbrace{P(\program|L)}_{\text{description-length prior}}
\end{equation}
For test images we relax the likelihood function to pixel-wise L2 distance; see Suppl.~Sect.~1.2.

The model learns from experience drawing. Given training images $\left\{I_n \right\}_{n = 1}^N$ the model updates its library $L$ by
searching to maximize:
\begin{equation}\label{libraryMap}
  P(L|\left\{I_n \right\}_{n = 1}^N)\propto \underbrace{P(L)}_{\text{description-length prior}}\times\prod_{n = 1}^N \sum_\program P(I_n|\program)P(\program|L)
\end{equation}

Equations \ref{programMap} and \ref{libraryMap} are intractable, because they require computing the infinite set of all possible programs. 
We approximate inference using an iterative approach based on the DreamCoder program synthesis algorithm (Figure \ref{algo}, Suppl.~Sect~1; see~\cite{dc2020,ecc}).
This alternates between inferring a program for each image (\textbf{Explore}),
updating the library $L$ with discovered subroutines used across program solutions (\textbf{Compress}), and training a neural network, $Q(\program|I)$, to predict a probability distribution over programs $\program$ likely to explain image $I$ (\textbf{Compile}). Learned subroutines can be abstract (e.g., taking as input arbitrary parameters or subprograms).

\paragraph{Converting programs to motor trajectories} 
Programs $\program$ are structural descriptions that do not represent ordering of strokes. For example, a program that translates a vertical line four times could correspond to either a left-to-right or a right-to-left drawing sequence. Thus, we ``ground'' these programs into possible motor trajectories $t$, defined for model and humans as an ordered list of segmented ``strokes'', each stroke summarized by a feature vector (see ``Analysis of behavior''). Each program $\program$ generates a set of \textit{admissible} $t$, which includes all trajectories whose stroke-sequence can be aligned with the program's syntax tree.

The probability of $t$ given program $\program$ in the program induction (PI) model is therefore given by:
\begin{equation}
  P(t|\program) = \frac{\indicator\left[t\text{ admissible for }\program \right]}{\sum_{t'}\indicator\left[t'\text{ admissible for }\program \right]}
\end{equation}

\begin{table}[ht]
\vspace{-0mm}
\begin{center} 
\caption{Starting primitives in library $L$. Model learns an inductive bias over programs by inducing new subroutines that are built out of these primitives.} 
\label{t:primitives} 
\begin{tabular}{llll} 
\hline
Primitive    & Arg. types\footnotemark & Returns & Description \\
\hline
line & none & $D$ & Line with endpoints at (0,0) and (1,0) \\
circle & none & $D$ & Unit circle centered at (0,0) \\
repeat  & ($D$, $n$, $T$) & $D$ & Drawing transformed by $T$ $n$ times \\
transform & ($D$, $T$) & $D$ & Applies affine transformation $T$ \\
reflect & ($D$, $\theta$) & $D$ & Reflects across axis defined by $\theta$  \\ 
connect & ($D$, $D$) & $D$ & Union of two drawings \\
affine & ($d$, $d$, $\theta$, $s$, $o$) & $T$ & Translation ($d$, $d$), rotation ($\theta$), scaling ($s$) in order $o$ \\
\hline
\end{tabular} 
\end{center} 
\vspace{-2mm}
\end{table}
\footnotetext{Types: $D$ is a drawing (set of ink coordinates); $N$, $\theta$, $d$, $s$, $o$ are discretized parameters drawn from a multinomial distribution} 

\paragraph{Reweighting motor trajectories by motor cost} 
Drawing is influenced by a variety of motor constraints \cite{goodnow1977children, forbus2011cogsketch, van1984drawing, lake2015human}. 
To bias the model toward efficient trajectories, we assign to each trajectory a score summarizing motor efficiency based on the statistics of human movement trajectories.

All possible motor trajectory permutations $t$ for a given program $p$ are assigned a motor cost. We define a \textit{feature extractor} $\phi(\cdot )$ that maps a trajectory $t$ to a trajectory-level real-valued feature vector $\phi(t)$ with four elements based \textit{a priori} on known drawing biases \cite{lake2015human, van1984drawing}: $start$ (position of first stroke relative to top-left corner), $distance$ (total movement distance), $direction$ (direction of movements relative to the diagonal), and $verticality$ (bias for vertical transitions) (see details in Suppl.~Sect 2.2).
Given an input image $I$, the model predicts a drawing trajectory $t$ with probability
\begin{equation}
    P(t|I) = \indicator\left[t~\text{draws}~I\right]\frac{\exp\left(-\theta\cdot\phi(t) \right)}{\sum_{t'}\indicator\left[t'~\text{draws}~I\right]\exp\left(-\theta\cdot\phi(t') \right)}\label{costModelDistribution}
\end{equation}

for weight-vector $\theta$, where $\theta\cdot \phi(t)$ is the cost of trajectory $t$.
Given a set of $N$ training images $\left\{I_n \right\}_{n = 1}^N$
%
and paired motor trajectories $\left\{t_n^{s} \right\}_{n = 1}^N$ for subject $s$, the model estimates for each subject 
$\theta^{s}$ via regularized maximum likelihood,
\begin{equation}
\theta^s = \argmin_{\theta^s}\sum_{n = 1}^N\underbrace{-\log P(t_n^{s}|I_n^{s})}_{\text{depends on }\theta^s\text{; Eq.}~\ref{costModelDistribution}} + \lambda||\theta^{s}||_2^2
\end{equation}
with a suitable coefficient of regularization $\lambda$. 


Finally, in the full model (i.e., ``Hybrid''; see below)  trajectories $t$ for program $p$ from the Program Induction model were reweighted using the Motor Cost model:
\begin{equation}
  P(t|\program,\theta^{gen}) = \frac{\indicator\left[t\text{ admissible for }\program \right]\exp(-\theta^{gen}\cdot \phi(t))}{\sum_{t'}\indicator\left[t'\text{ admissible for }\program \right]\exp(-\theta^{gen}\cdot \phi(t'))}
\end{equation}
Where $\theta^{gen}=\frac{1}{S}\sum_{s=1}^{S}\theta^s$ is a set of \textit{general} parameters averaged across all subjects.



\textbf{Lesioned models}
The full model (HM) was compared to ``lesioned'' models in Table~\ref{t:models}.


\begin{table}[ht]
\vspace{-0mm}
\begin{center} 
\caption{Models used in this study} 
\label{t:models} 

\begin{tabular}{llll} 
\hline
Model    & Training set &  Abbrev. & Description \\
\hline
Null & none & Null & All trajectories, equal probability \\
Motor Cost & 1, 2 (motor data) &  MC1, MC2 & All traj., w/group-specific cost \\
Program Induction  & 1, 2 (images) &  PI1, PI2 & Admissible traj., all equal probability \\
Hybrid & 1, 2 (mot. \& im.) & HM1, HM2 & Admissible traj., w/across-group cost \\
\hline
\end{tabular} 
\end{center} 
\vspace{-2mm}
\end{table}

\begin{figure}[ht]
\begin{center}
    \includegraphics[width=1.0\linewidth]{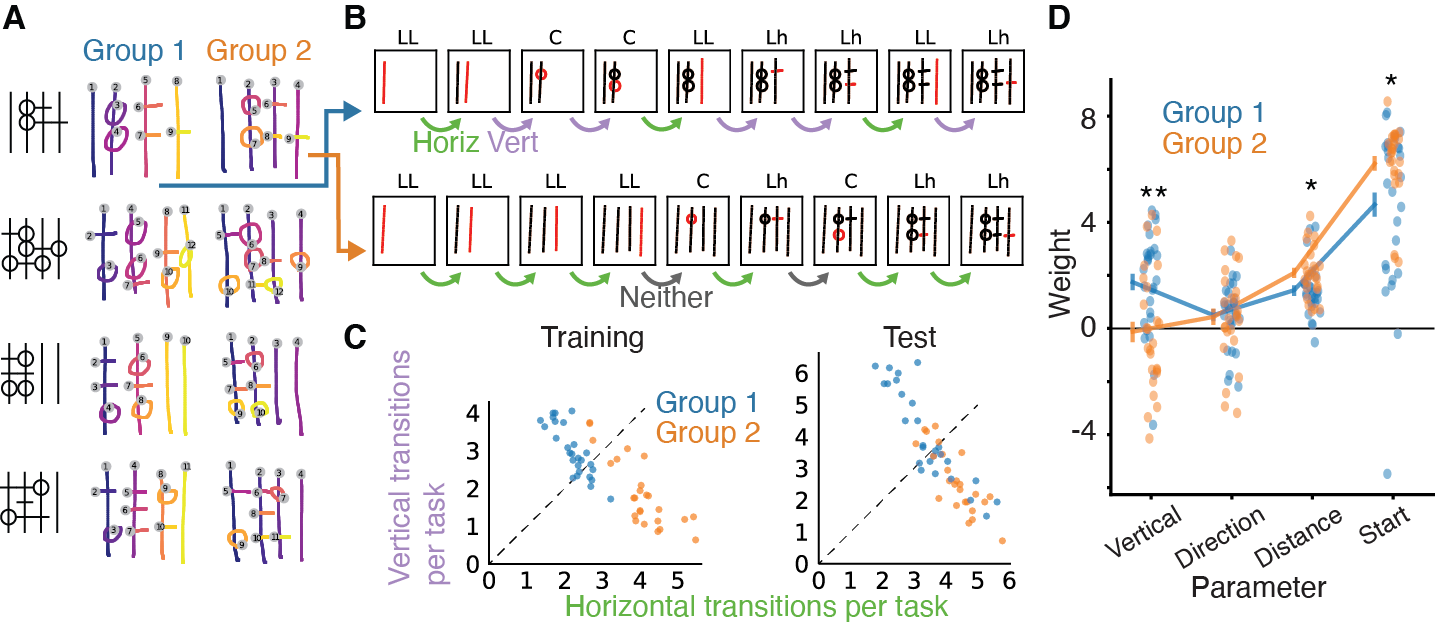}
    \caption{ Single-session learning of structure in drawings. \textbf{(A)} Example drawing trajectories for two subjects (columns) on four test tasks (rows). Stroke order is indicated by both color (purple to yellow) and the number in grey dot. Grey dots also indicate start positions. \textbf{(B)} Example segmentations. Letter codes indicate stroke categories. \textbf{(C)} Stroke transitions were directionally biased. \textbf{(D)} MC feature weights for behavior on Test stimuli. Positive corresponds to a bias for transitions that are vertical, towards bottom-right, low distance, and for first stroke at top-left. *, **, p$<$.05, .005, t-test.}
    \label{behavior}
\end{center}
\vspace{-6mm}
\end{figure}

\section{Experiments}

\subsection{Methods}


\paragraph{The drawing task}
Humans and models copy visually-presented objects defined by simple objects and composed geometric rules (Figure \ref{background}). Humans were randomly split into two groups whose training sets differed in higher-order structure. Generalization was tested on a single common set of tasks. Therefore, differences in behavior on test stimuli are attributable to learning on the training sets. 

\paragraph{Subjects}
Subjects [N = 104 (58 M, 44 F, 2 excluded due to incorrect copying of the objects or errors in saving data), Age = 35.0 +/- 9.3 (mean/SD)] were recruited on Amazon Mechanical Turk and paid \$3.00 for 15-20 minutes. Subjects gave informed consent. The study was approved by our institution's Institutional Review Board.

\paragraph{Stimuli} See examples in Figure~\ref{background}C. Two sets of stimuli were generated using different probabilistic algorithms. Both had repeated vertical lines (2, 3, or 4), but differed in the other components. Set 1 had vertically grouped strokes (lines and circles) superimposed on the vertical lines, while Set 2 had horizontally-oriented groups of strokes, sampled from an library of objects (e.g. dumbbells (o--o), lollipops (--o) and poles (---)). We randomly generated 250 stimuli for each training set, from which we manually selected 36 representative samples. The common Test set included 18 manually designed ambiguous images.

\paragraph{Procedure} The experiment was presented in a web browser using PsiTurk \cite{gureckis2016psiturk} on a touchscreen device (phone, tablet, or laptops). The instructions read: \textit{You will learn to write letters from an alphabet of an alien civilization recently discovered by astronauts. Scientists would like to study how people learn to write new alphabets. Your task is to copy the letters. Try to be quick, but it is also important to be accurate! Letters are taken from the same alphabet. But letters get harder over time, so try to learn from the earlier trials!}
On each trial a single stimulus was presented top-center of the screen. The subjects copied it on ``sketchpad'' directly below the stimulus without explicit time constraints or evaluative feedback. Subjects first saw seven simple stimuli (e.g the first three stimuli in Figure ~\ref{background}C), followed by 13  training stimuli of varying difficulty. Next, subjects copied the 18 testing intermixed with the remaining 16 training stimuli, in orders randomized for each subject.

\paragraph{Analysis of behavior}
Motor trajectories were segmented into discrete \textit{strokes}, and each stroke was summarized by a feature vector $\phi_{stroke}$ = ($category$, $startLocation$, $center$, $row$, $column$) (details in Suppl.~Sect 2.1). Each trajectory was defined by an ordered list of strokes: $(\phi_{stroke}^1, \phi_{stroke}^2, ...)$.



\paragraph{Scoring model-human distance} For each combination of test image $I$, human $h$, and model $m$, we measured the distance between human behavior and model predictions by: 
\begin{equation}
    d(h, m, I) = \sum_{t'}\text{Damerau–Levenshtein}(t', t^{h})P(t'|\program^{m},\theta^{m})
\end{equation}
where the Damerau–Levenshtein edit distance is applied to pairs of trajectories, $p$ is the highest-scoring program based on the program induction model, and $\theta$ are motor cost parameters.

\subsection{Human results}



\paragraph{Rapid learning of structure in drawings}
We expected that a behavioral readout of learning would be for subjects trained on Task 2 (horizontally structured objects) to produce a relatively higher frequency of horizontal transitions, compared to subjects trained on Task 1 (vertically structured objects) (Figure~\ref{background}C). Indeed, we found this to be the case (Figure~\ref{behavior}A-C).
We confirmed that this difference in vertical vs. horizontal biases remained even after accounting for other changes to behavior. We jointly fit parameters describing four different motor features: \textit{Start, Distance, Direction}, and \emph{Verticality}, fit separately for each subject using the Motor Cost model. 
While we found weak, but significant, differences in \textit{Start} and \textit{Distance} between the two groups, we found a relatively strong difference in \textit{Verticality} weights, consistent with the previous analysis of transition frequencies (Figure \ref{behavior}D). This result further supports that training led to strong apparent biases for vertical vs. horizontal transitions.

\begin{figure}[htb]
\begin{center}
    \includegraphics[width=0.95\linewidth]{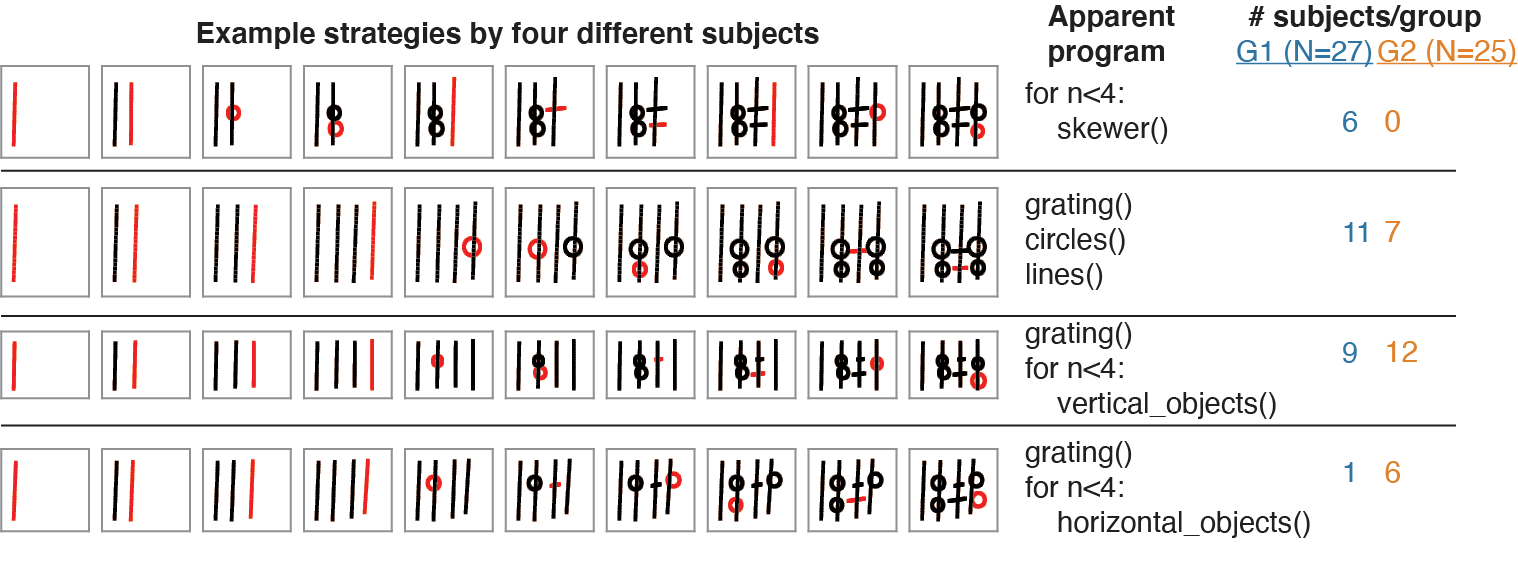}
    \caption{ Program-like structure in behavior. Four example subjects depicting each strategy (left), apparent program-like structure (middle), and frequencies of these strategies for the two training groups.}
    \label{strategies}
\end{center}
\vspace{-4mm}
\end{figure}

\paragraph{Program-like structure in behavior} 
Did differences in vertical and horizontal biases reflect changes in lower-level motor preferences, or more abstract biases? Consistent with abstraction, subjects' qualitative behavior appeared to be well-described by abstract programs (Figure ~\ref{strategies}). In particular, four program-like strategies were prominent across subjects. The ``skewers'' strategy involved drawing a vertical line, immediately followed by the objects ``skewered onto it'' (Figure~\ref{strategies}, top row). ``Skewers'' was used only by Group 1 subjects. Other strategies involved first drawing the vertical gratings followed by different ways of drawing the smaller objects (Figure~\ref{strategies}, rows 2-4).
We quantitatively assigned one strategy to each subject based on the distribution of Motor Cost model parameters, extended with two additional parameters capturing biases for (1) perseverating on a given category of objects (e.g., circle-circle-circle...), and (2) finishing a ``skewer'' before before moving to the next one (see Suppl.~Sect 2.3). Group 1 and Group 2 subjects tended to use different program-like strategies (Figure~\ref{strategies}, right).

\begin{figure}[ht]
    \begin{minipage}[c]{0.50\textwidth}
    \includegraphics[width=\linewidth]{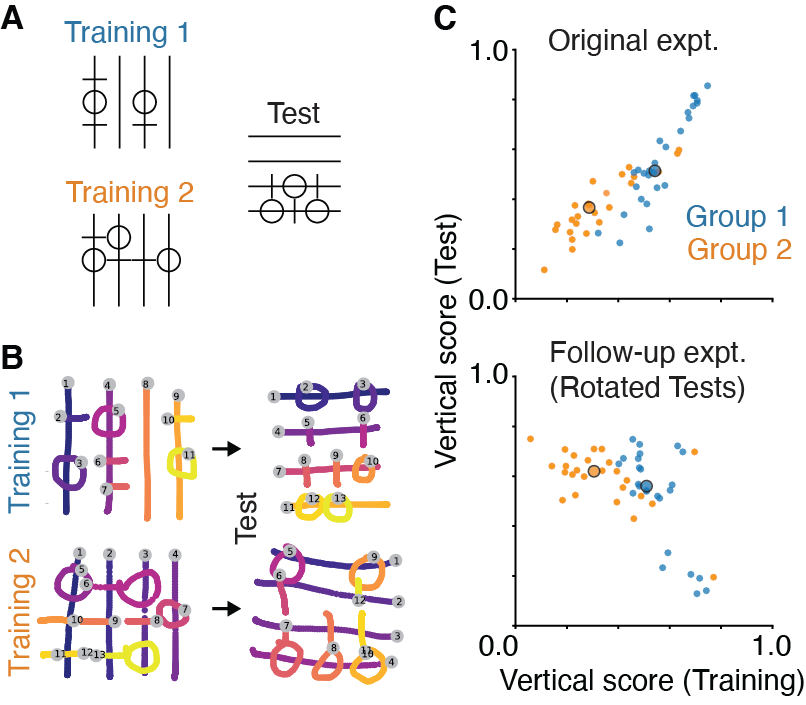}
    \end{minipage}\hfill
    \begin{minipage}[c]{0.35\textwidth}
      \caption{Evidence for abstract generalization in a followup study with rotated test stimuli. \textbf{(A)} Test stimuli are rotated relative to the original experiment in Figure~\ref{behavior}. \textbf{(B)} Example drawings for two subjects showing that they retained program-like biases evident for rotated Test tasks. \textbf{(C)} Summary analysis. ``Vertical score'' is computed as V/(V + H), where V and H are average vertical and horizontal transitions per task. Large dots indicate  medians.}
      \label{rotation}
    \end{minipage}
\vspace{-4mm}
\end{figure}

\paragraph{Evidence for abstract generalization} 
On a new set of subjects, we performed a modified experiment to further test whether subjects indeed learned abstract programs. We reasoned that abstract programs should persist if the Test stimuli were rotated (Figure~\ref{rotation}A). However, if subjects learn only the horizontal vs. vertical motor biases, we would expect this bias to remain unchanged. We found that directional biases changed orientation when the test stimuli were rotated (Figure~\ref{rotation}B,C). In the original experiment subjects in Group 1 exhibited a stronger vertical bias than those in Group 2 during Training, and this effect carried over to Testing (replotted in Figure~\ref{rotation}C). However, in the modified experiment, while Group 1 subjects still exhibited a stronger vertical bias during training, they preferred horizontal transitions when tested on rotated stimuli (Figure~\ref{rotation}C). This flexible adaptation of directional biases in a manner that mimics the orientation of the stimuli is consistent with the learning of abstract programs.

\subsection{Modeling results}

\begin{figure}[ht]
\begin{center}
    \includegraphics[width=1.0\linewidth]{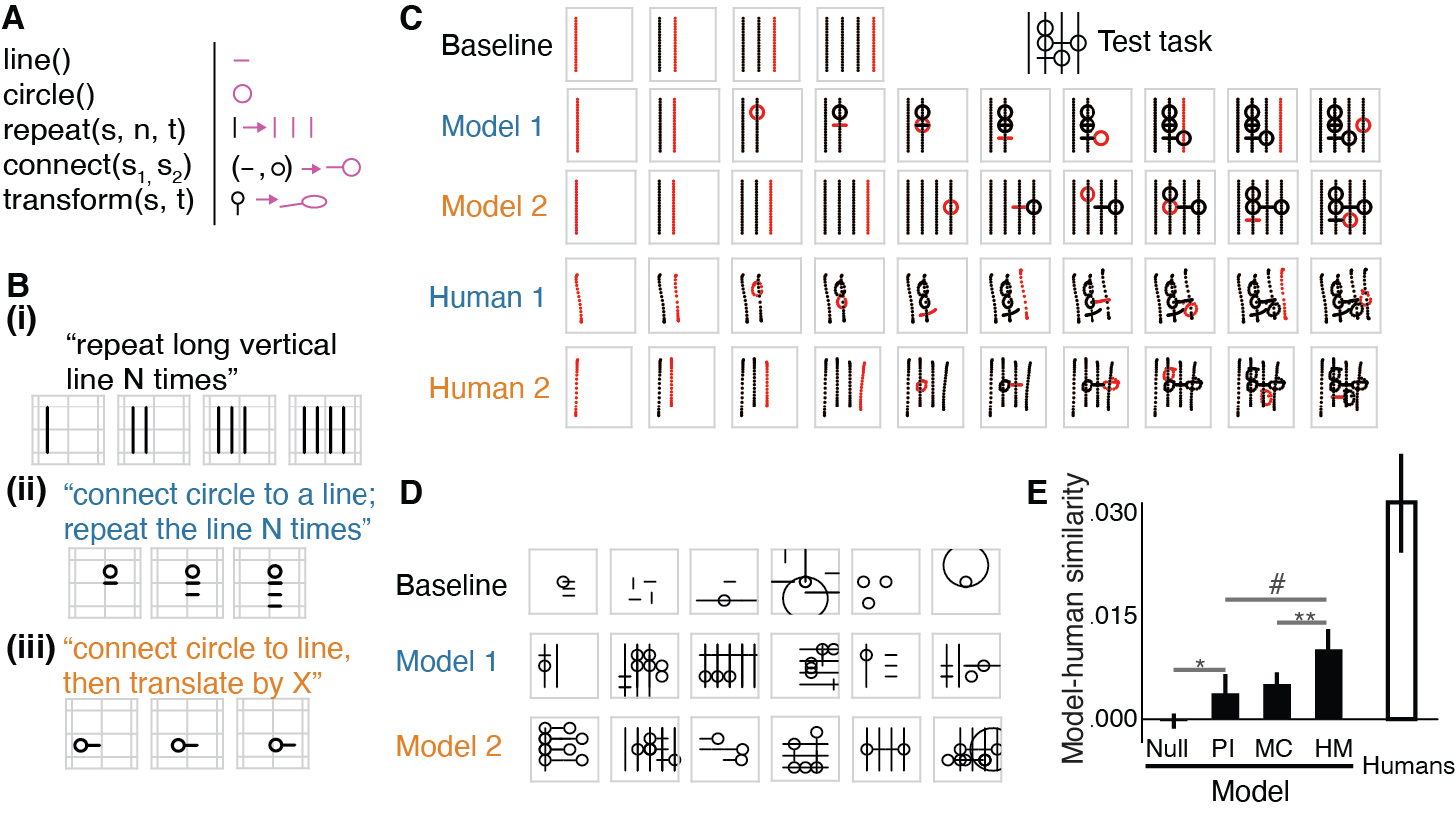}
    \caption{Modeling results. \textbf{(A)} Example starting primitives (left) and drawings (right). \textit{s, n} and \textit{t} are variables representing drawings, natural number, and transformation (see Methods). \textbf{(B)} Example subroutines learned by both PI1 and PI2 (Bi), PI1 only (Bii), and PI2 only (Biii). \textbf{(C)} On an example test task, solutions by models (top) and example humans (bottom). \textbf{(D)} Example ``dreams,'' or prior samples at Baseline, and after training on Tasks 1 (Model 1) or 2 (Model 2).\textbf{(E)} Comparing human and model behavior on test tasks. $Similarity = mean(D(H1, M2), D(H2, M1)) - mean(D(H1, M1), D(H2, M2))$, where $D(H', M')=\frac{1}{N_h}\frac{1}{N_s}\sum_{h \in H'}\sum_{s}dist(h, M', I_s)$ is distance to model averaged over humans ($h$) and test stimuli ($s$). *, **, p$<$.05, .005; \#, p=.06, paired t-test. Null model p$<$.05 vs. other models.} 
    \label{dreamcoder1}
\end{center}
\vspace{-6mm}
\end{figure}

\textbf{Program induction}
We trained a pair of models on either Training sets 1 (HM1) or 2 (HM2),
initialized with the primitives in Table~\ref{t:models} (some drawn in Figure~\ref{dreamcoder1}A). 
The models successfully learned new program subroutines (examples in Figure~\ref{dreamcoder1}B). This was reflected in unconditioned samples from learned priors (i.e., ``dreams''), which exhibited different, task-related structure, such as vertical ``skewers'' for Model 1 and horizontal ``barbells'' for Model 2. Some dreams also extrapolated from the training stimuli (Figure~\ref{dreamcoder1}D). Untrained models did not exhibit task-related dreams (Figure~\ref{dreamcoder1}D, ``Baseline''). Trained models performed well on the test tasks [mean/SD of 1.9/1.7 (HM1) and 0.35/0.76 (HM2) mistakes (missed or extra strokes) per task (out of 11.6)]; in contrast, the untrained model was unable to solve the test tasks. Moreover, the two trained models often produced different solutions to the same task, coarsely resembling solutions produced by trained humans (Figure~\ref{dreamcoder1}C). 

\textbf{Comparison with humans} We quantified the similarity between humans and models, finding that Human group 1 was better fit by HM1 than by HM2, and Human group 2 was better fit by HM2 than by HM1, indicating that learning altered the structure of behavior for both humans and models in overlapping ways (Figure~\ref{dreamcoder1}E). We also compared HM1 and HM2 to alternative ``lesioned'' models (see Methods and Table~\ref{t:models}). HM modeled humans better than did Program Induction alone (PI), consistent with significant contribution of motor biases. Moreover, HM performed better than a model with learned motor biases but without program induction (MC), indicating an important role of abstract program-like structure in learning. As an upper bound, we assessed the similarity of humans to other humans in the same Training group (Figure~\ref{dreamcoder1}E). Not surprisingly, compared to this upper bound, the HM model did not capture all complexities of human behavior. This is partly due to human drawing being influenced by a variety of biases which we did not attempt to model (see Discussion). This result argues that a model like ours, trained on a small dataset and without access to human data, captures diagnostic features of human generalization. 



\section{Discussion}
We show that humans learn new abstract program-like structure from brief training on drawing tasks, and with no explicit instruction. To understand this learning computationally, we built a generative model combined with a learning algorithm formalizing the principles of abstraction, compositionality, and motor efficiency.
Trained on the same data as humans, this model learned a new set of abstract drawing subroutines by recombining a small set of simple drawing primitives. These learned subroutines support generalization behavior that resembles important aspects of human drawings in this task. Our results suggest that principles of abstraction and compositionality are central to explaining how humans learn generalizable program-like structure in drawing. 


We formalize learning as acquisition of parsimonious internal models that explain shared structure underlying multiple learning experiences. The idea of knowledge as efficient abstraction has parallels in philosophy (e.g. \textit{Occam's razor}), psychology~\cite{feldman2016simplicity}), and cognitive modeling (e.g. hierarchical Bayesian accounts of \textit{learning-to-learn} \cite{kemp2007learning, lake2015human, tenenbaum2011grow} ). 
Our computational approach to learning builds on this work, 
by representing concepts as generative programs, and inductive biases as priors over programs~\cite{fodor1975language, goodman2008rational, cheyette2017knowledge, rule2018learning, lake2015human}. 


One limitation of our study is that stimuli were relatively simple and ``clean.'' This was by design, as we focus less on real-world writing or drawing skill (e.g., \cite{lake2015human}) but on abstraction in rapid learning. Second, we did not attempt to capture the full complexity of drawing, which likely contributes to the diversity of behavior across people. Future work may attempt to model this diversity as differences in starting priors (e.g., related to perception, motor skill, art and writing experience, and others).

Recent deep-learning-based models have had success modeling drawing and handwriting on more complex concepts than considered here. However, in contrast to our model, they need significantly larger training data-sets, in some cases supervision with human motor behavior \cite{ha2017neural, zhang2017drawing}, 
and are usually not systematically compared to how humans learn new inductive biases ~\cite{ha2017neural, mellor2019unsupervised, gregor2015draw, zhang2017drawing, zhou2018learning}. The speed of human learning in our task, paralleled by the model, highlights the importance of learning rich and flexible structured representations for generalization, either explicitly~\cite{lake2015human,mao2019neuro} or implicitly, as in differentiable neural computers~\cite{graves2016hybrid} and others~\cite{reed2015neural}. Similar to children acquiring sophisticated knowledge in a manner bootstrapped by ``core'' systems of knowledge~\cite{spelke2017core,carey2009origin,huang2015core}, our study supports the view that the rapid learning of structured representations can result from appropriate starting primitives coupled to learning algorithms guided by abstraction and compositionality.

\section{Broader Impact}
We envision a number of scientific, societal and engineering benefits that may emerge from this study. First, this work may benefit the treatment, diagnostics, and prevention of cognitive disorders, such as disorders involving planning, reasoning, and learning. The methodology of our task is particularly relevant for understand disorders that cause striking impairment in drawing behavior (for example, dementia, traumatic brain injury, and stroke). A computational understanding of cognitive impairment may lead to more accurate, quantitative diagnostic tools (by categorizing disorders based on cognitive computations) and to more efficient targeted treatment (by targeting of specific impairments).

Second, computational understanding of how humans think is scientifically worthwhile, because it advances our understanding of nature and the human condition. In addition to the current study of human adults, we are studying this task in human children, and in non-human primates in a neurophysiological setting, with the goal of also studying this task at a neural level. As the long-term goal of this multi-species investigation we hope to develop an evolutionary, developmental and longitudinal understanding of learning of complex structure, as emerging from simple components and basic computational principles. 

Third, from an engineering standpoint, this work may lead to AI that is more easily integrated into, and more beneficial to society. The ability to learn new human-like inductive biases from a small number of examples may facilitate human-computer interaction, particularly within programming-by-examples technologies~\cite{gulwani2011automating}.
Engineering outcomes of this research may also contribute to tools that benefit education. The link to learning drawing and art is obvious, but there may also exist links to topics that involve structured symbolic reasoning, such as math, science, or music. For instance, modeling a given student's learning trajectory may reveal what she knows and what strategies she uses to learn, which may suggest ways to either tailor her future learning, or to remedy current difficulties. Indeed, we also plan to study variants of this task in children.

In principle, work along this line may potentially be used to create ``fake'' artifacts meant to pass as human. The most obvious kinds of fake artifacts are those related to drawing, but this extends to other kinds of art and media, such as internet bots that impersonate humans by generating tweets from examples. One potential implication is that fakes will be used to confuse and manipulate society. Ways to address this should fall under strategies and considerations already being developed to understand the impact of deep-fakes on society. A second implication is that mass-produced AI artifacts could lower the quality of creative content in the world, and compete with high-quality human-made creative products. We think of this possibility as an ethical uncertainty, and note that it is an extension of the apparent already-occurring trend towards larger amounts of mass-produced media in society.

\bibliographystyle{unsrt}
{\small \bibliography{main}}

\end{document}


\maketitle

\section{Program synthesis algorithm}
The algorithmic engine behind our program synthesis method follows the approach introduced in EC$^2$~\cite{ecc}, and is based on the open source implementation of EC$^2$'s successor, DreamCoder~\cite{dc2020}.
Our program induction model takes as input a corpus of black-and-white raster \textbf{training images}, and seeks to synthesize a graphics program that generates each of them. The model estimates a prior over programs for training images, to be deployed on held-out \textbf{test images}.
Following~\cite{dc2020} we now derive this algorithm starting from a Bayesian viewpoint.
With this probabilistic formalism in hand we will then briefly outline the 3-step algorithm which performs inference in this model, but interested readers should consult~\cite{dc2020,ecc} for a complete algorithmic exposition.

\noindent\textbf{Notation.} We write $I$ to mean an image, and $\rho$ to mean a graphics program.
Programs are represented in typed lambda calculus.
Initially the graphics programming language contains the \textbf{primitives} outlined in Table~1 of the main text. Primitives are expressions in typed lambda calculus.

We write $\denotation{\rho}$ to mean the image output by program $\program$.
We write $L$ to mean a \textbf{library} of primitives;
At the initial state of learning $L$ contains the primitives in Table~1 of the main text.
The library $L$ acts as a prior over the space of programs, written $P(\cdot |L)$ and defined formally in~\cite{dc2020}.
Intuitively, this prior prefers programs which may be expressed compactly using the primitives in $L$.
We write $P(L)$ to mean the prior probability of the library $L$, and this prior prefers libraries which overall contain less code (smaller lambda calculus expressions).

From a Bayesian point of view our aim is to estimate the prior maximizing the joint probability, which we will notate $J$:
\begin{equation}\label{libraryMap}
  P(L|\left\{I_n \right\}_{n = 1}^N)\propto P(L,\left\{I_n \right\}_{n = 1}^N) =  J(L) = P(L)\prod_{n = 1}^N \sum_\program P(I_n|\program)P(\program|L)
\end{equation}
where $P(I|\program) = \indicator\left[I = \denotation{\program} \right]$.
Evaluating this objective is intractable because it requires marginalizing over the infinite set of all programs.
We define the following intuitive lower bound, written $\lowerBound$, on this objective function:
\begin{equation}
  J(L)\geq \lowerBound(L,\left\{\mathcal{B}_{I_n} \right\}_{n = 1}^N) = P(L)\prod_{1\leq n\leq N}\sum_{\program\in \mathcal{B}_{I_n}}P(I_n|\program)P(\program|L)
\end{equation}
where the bound $\lowerBound$ is expressed in terms of a collection of sets of programs, $\left\{\mathcal{B}_n \right\}_{n = 1}^N$, called \textbf{beams}:
\\\noindent\textbf{Definition.} A \textbf{beam} for image $I$ is a finite set of programs where, for any $\program\in \mathcal{B}_I$, we have $P(I|\program) > 0$.
In other words, every program in the beam for image $I$ correctly draws $I$.

Making the beams finite ensures that calculation of $\lowerBound$ is tractable.
In our experiments we bounded the size of the beams to 5.

We alternate maximization of $\lowerBound$ with respect to the beams and the library.
In reference to figure~2 of the main text, these alternate maximization steps are called the \textbf{Explore} and \textbf{Compress} steps.
\\\noindent \textbf{Explore: Maxing $\lowerBound$ w.r.t.\ the beams.} Here $L$ is fixed and we
want to find new programs to add to  the beams so that $\lowerBound$ increases the most.
$\lowerBound$ most increases by finding programs where $\probability[I|\program ]\probability[\program|L]$
is largest.
\\\noindent \textbf{Compress: Maxing $ \lowerBound$ w.r.t.\ the library.} Here $\left\{\mathcal{B}_{I_n} \right\}_{n = 1}^N$ is held fixed, and so we can evaluate $\lowerBound$. Now the problem is that of searching the discrete space of libraries and finding one maximizing $\lowerBound$.

Searching for programs is hard because
of the large combinatorial search space. We ease this difficulty by training a neural recognition model, $Q(\cdot |\cdot )$,
during the Compile step: $Q$ is trained to approximate the
posterior over programs, $Q(\program|I)\approx P(\program|I,L)$,
  thus amortizing the cost of finding programs with high posterior probability.

\textbf{Compile: learning to tractably maximize $\lowerBound$ w.r.t. the
  beams.}  Here we train 
$Q(\program|I)$ to assign high probability to programs $\program$ where
$P(x|p)P(p|L)$ is large, because including those programs
in the beams will most increase $\lowerBound$.
We train $Q$ both on programs found during the Explore step and
on samples from the current library, i.e. $P(\cdot |L)$.
Assuming that $Q$ successfully converges to the true posterior estimates, then
incorporating these top programs as measured by $Q$ into the beams
will maximally increase $\lowerBound$.

\subsection{Algorithmic details}

Having introduced the probabilistic framing of our problem, and the 3-step inference procedure, we now briefly outline the algorithmic implementation of the explore/compress/compile steps.
A full overview is contained in~\cite{dc2020}.

\subsubsection{Explore}

During the Explore step we enumerate programs in decreasing order under  $Q(\cdot |I_n)$ for each image $I_n$, and keep the top 5 within the beam $\mathcal{B}_{I_n}$ as measured by the posterior $P(\program|I_n,L)$.
This enumeration is tractable given our parameterization of $Q$, which can be expressed similarly to a PCFG over programs;
i.e.,  the neural network outputs a distribution over programs parameterized by the weights of a probabilistic grammar, and we enumerate in decreasing order under that grammar.
Thus, $Q$ outputs the transition probabilities of a bigram model over program syntax trees, which may be unrolled into a PCFG-like representation.
Enumeration proceeds until a per-image timeout is reached; we used a timeout of one hour.

\subsubsection{Compress}

Here we seek to update the library by increasing the probability it assigns to programs in the beams, hence ``compressing'' those programs. Indeed the compress objective can be rewritten in terms of a sum of description lengths:
\begin{equation}
  \argmax_L \lowerBound = \argmin_L\underbrace{-\log P(L)}_{\text{description length of library}} + \sum_{1\leq n\leq N}\underbrace{-\log \sum_{\program\in \mathcal{B}_{I_n}}P(I_n|\program)P(\program|L)}_{\text{description length of program generating image }I_n}
\end{equation}
To heuristically minimize this description length we search locally through the space of libraries $L$ until the above objective fails to improve.
Our search moves consist of incorporating new subexpressions obtained from automatically refactoring programs in the beams---intuitively, refactoring programs that we found explaining images so as to minimize the total size of the library plus the total size of those programs.
This refactoring process combines version space algebra~\cite{lau2003programming} with equivalence graphs~\cite{tate2009equality}, which are two approaches from the programming languages and program synthesis community; see~\cite{dc2020} for details.

\subsubsection{Compile}

Here we train a neural network to guide the search over programs, seeking to minimize its divergence from the true posteriors over programs. Writing $\phi$ for the parameters of $Q$, we aim to maximize
\begin{equation}
  \argmax_\phi\Expect\left[Q_\phi\left( \left(\argmax_\program P(\program|L,I) \right) \;\big\vert\;I \right)\right]
\end{equation}
where the expectation is taken over images $I$.
Taking this expectation over the empirical distribution of images trains the network on programs found during the Explore step; taking the expectation over samples, or ``dreams,'' from the learned prior $L$.
Training on dreams is critical for sample efficiency: just like humans our model learns from at most a few dozen images, which is too little training data for a high-capacity neural network.
But as we learn our prior, we can then draw unlimited dreams to train the neural network.

\subsection{Generalizing to test images}

Prior to evaluation on test images we iterate this learning procedure for 20 cycles (of searching for task solutions, updating the library, and training the neural network).
The end state of learning is not just a program for each training image but, critically, also a learned inductive bias $L$ and a learned inference/synthesis strategy $Q$.
When comparing with human data we infer a program for test images by enumerating programs in decreasing order under $Q(\cdot |I)$ and then rescoring under $P_\text{test}(I|\program)P(\program|L)$,
where the likelihood $P_{\text{test}}(I|\program) \propto\exp \left(-|I - \denotation{\program}|_2 \right)$ is a relaxed version of the 0/1 likelihood during train to allow partial credit when the model cannot fully explain a test image.

\section{Analysis of behavior}

\subsection{Converting a motor trajectory into a sequence of discrete strokes}
Motor trajectories [raw data in the form of coordinates and corresponding times $(x, y, t)$] were segmented into discrete ``strokes''. Each stroke was a sequence of coordinates during which the finger was continuously touching the screen; i.e., strokes were separated by no-touch gaps. Subjects naturally tended to lift their finger between each discrete segment (i.e., line or circle) in the drawing. 

Each stroke was summarized in a stroke-level feature vector $\phi_{stroke}$ = ($category$, $startLocation$, $center$, $row$, $column$). $category$ is the type of object represented by the stroke, either a ``vertical line'' (LL), ``horizontal line'' (Lh), or ``circle'' (C). $startLocation$ is the $(x, y)$ position of the stroke onset. $row$ and $column$ were defined by "snapping" a stroke onto its position in a $3 \times N$ ($\text{row}\times \text{column}$) grid, where $N$ is the number of vertical lines in the ``grating'' for a given stimulus. 

An entire trajectory $t$ was therefore defined by the ordered list of strokes: $(\phi_{stroke}^1, \phi_{stroke}^2, ...)$.

To calculate frequencies of horizontal/vertical transitions in a given trajectory, stroke transitions were classified as either horizontal (same row, different column), vertical (same column), or undefined (different row, different column).

\subsection{Motor Cost model: extracting motor cost features from motor trajectories}

We define a \textit{feature extractor} $\phi(\cdot )$ that maps a trajectory $t$ to a trajectory-level real-valued feature vector $\phi(t)$ with four elements based \textit{a priori} on known drawing biases \cite{lake2015human, van1984drawing}: $start$, $distance$, $direction$, and one meant to capture biases reflecting learning in this task $verticality$.

$start$ is position of first touch, represented as distance from the top-left corner. $distance$ is the summed distance traveled, measured as the path between stroke centers, $direction$ is the summed distance travelled, but projected onto the diagonal from top-left to bottom-right, a direction chosen because it is a common bias in sketching and writing. $verticality$ is the cumulative distance moved projected onto the y-axis subtracting distance projected onto the x-axis, and was included to allow the Motor Cost models trained on separate groups (MC1 and MC2) to capture learned directional biases. 

\subsection{Extending motor cost features to quantify program-like structure in behavior}

In order to quantify variation in behavior across subjects, we fit the Motor Cost model to each subject, but with the model extended with two parameters in addition to the four described above, $chunking$ and $skewers$. $chunking$ was the count of the number of transitions between strokes of identical categories (e.g., circle -> circle), which reflects a bias to group similar objects together. $skewers$ quantifies whether transitions away from ``long vertical line'' tend to be vertical (reflecting a bias to draw skewers) or horizontal (reflecting a bias to draw gratings); this was implemented similarly to $verticality$.

Subjects exhibited program-like structure as described in Figure 4 in the main text. Assignment of subjects to different strategies was supported both by visual inspection of their behavior and of their features from fitting this extended Motor Cost model (Supplemental Figure 1). We found the outcomes of these two methods to be largely in agreement. We note that subjects did tend to form a continuum between these four strategies, and so we attempted to assign the dominant strategy. 

\begin{figure}[htb]
\begin{center}
    \includegraphics[width=0.9\linewidth]{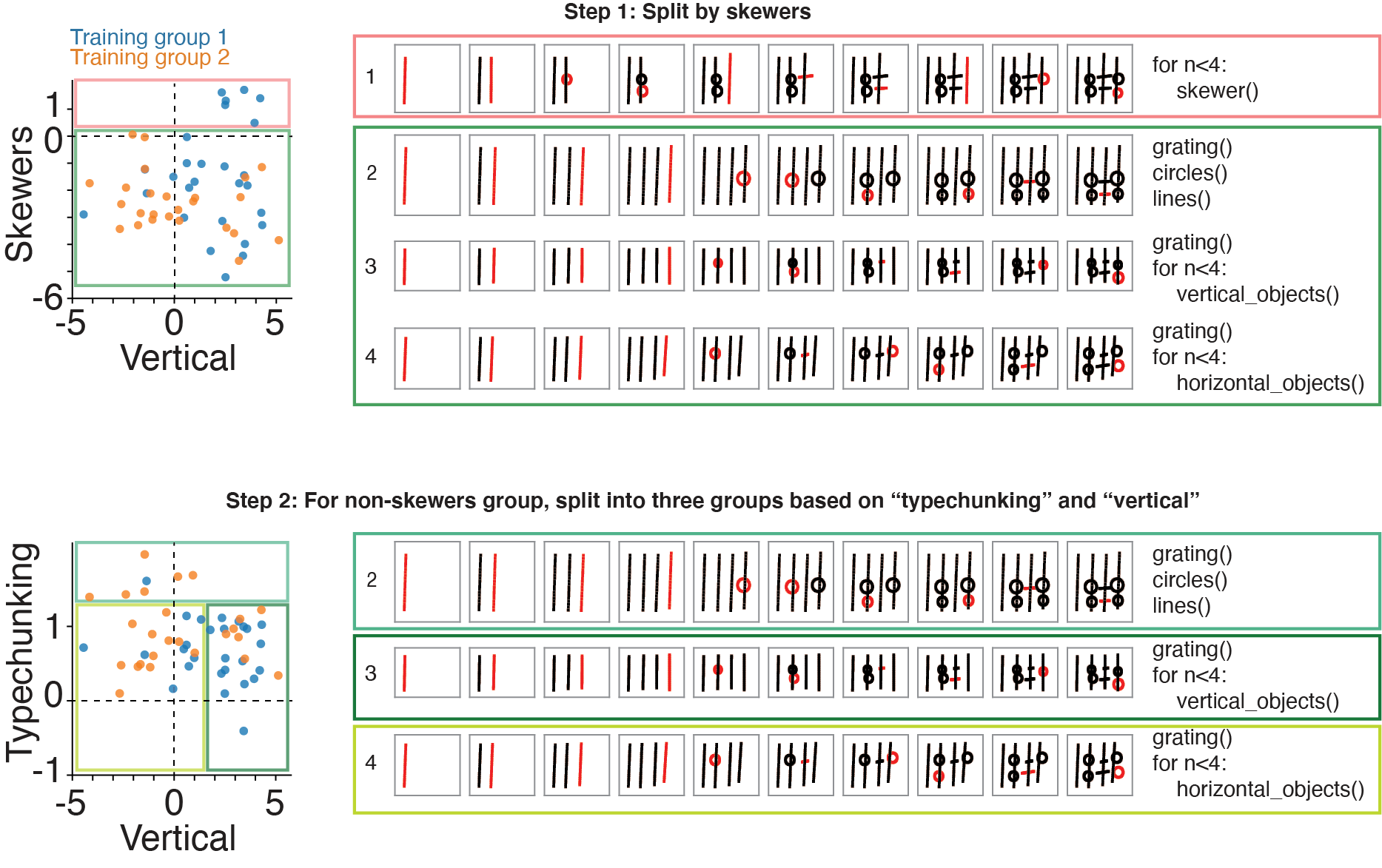}
    \caption{ Quantitation of diversity of program-like strategies. Left scatter plots: datapoints represent values of Motor Cost model features for individual subjects along three dimensions: ``skewers'', ``vertical'', and ``typechunking'' are Motor Cost model features. Top row: first, subjects were grouped into those drawing ``skewers'' and those drawing ``gratings'' (Step 1). Bottom row: second, the ``gratings'' group was split into three groups based on what they drew immediately after drawing gratings (Step 2). These two steps led to four mutually exclusive groups. Colored boxes allow for comparison between the scatterplots (left) and example motor sequences (right).} 
    \label{background}
\end{center}
\vspace{-6mm}
\end{figure}

\section{Datasets}
Included are behavioral data from all experiments in this paper. 
\paragraph{Supplementary\_dataset\_1.pickle}
Data from 54 subjects in the main experiment.
\paragraph{Supplementary\_dataset\_2.pickle}
Data from 50 subjects in the followup experiment (rotated test stimuli).




\bibliographystyle{unsrt}
{\small \bibliography{main}}